\documentclass{article} 
\usepackage{iclr2027_conference,times}
\usepackage{graphicx}
\usepackage{subcaption}
\usepackage{amssymb}
\usepackage{placeins}
\usepackage{float}
\usepackage[ruled,vlined,linesnumbered]{algorithm2e}
\usepackage{booktabs}
\usepackage{multirow}
\usepackage[table]{xcolor}
\usepackage{wrapfig}
\definecolor{oursblue}{RGB}{226,238,249}
\PassOptionsToPackage{pagebackref}{hyperref}
\newcommand{\ourscell}[1]{%
    \cellcolor{oursblue}#1%
}
\usepackage{makecell}

\newcommand{\mainTableDash}{%
  \noalign{\vskip 1pt}%
  \multispan{11}%
  \leaders\hbox{\rule{2pt}{0.4pt}\kern 2pt}\hfill
  \cr
  \noalign{\vskip 1pt}%
}

\DeclareRobustCommand{\pdfdiff}[1]{{\color{black}#1}}

\usepackage{amsmath,amsfonts,bm}

\def\eqref#1{equation~\ref{#1}}

\def\1{\bm{1}}

\DeclareMathAlphabet{\mathsfit}{\encodingdefault}{\sfdefault}{m}{sl}
\SetMathAlphabet{\mathsfit}{bold}{\encodingdefault}{\sfdefault}{bx}{n}

\usepackage{hyperref}
\usepackage{url}
\usepackage{booktabs}
\usepackage{array}
\usepackage{tabularx}

\title{Dynamic Model Merging with Multi-Layer Aggregated Routing Evidence}

\author{%
\begin{minipage}[t]{\dimexpr\textwidth-2\tabcolsep\relax}
\centering
\normalfont
\textbf{Yuxi Li}\textsuperscript{1},
\textbf{Jiacong Hu}\textsuperscript{2},
\textbf{Jie Song}\textsuperscript{1,2},
\textbf{Mingli Song}\textsuperscript{1,2},
\textbf{Yuanyu Wan}\textsuperscript{1,2}
\\[5pt]
\textsuperscript{1}School of Software Technology, Zhejiang University
\\
\textsuperscript{2}State Key Laboratory of Blockchain and Security, Zhejiang University
\\[5pt]
\texttt{\{yuxili,jiacong,sjie,songml,wanyy\}@zju.edu.cn}
\end{minipage}%
}

\iclrfinalcopy 
\begin{document}

\maketitle
\fancyhead{}
\renewcommand{\headrulewidth}{0pt}

\begin{abstract}
Dynamic model merging aims to retain the efficiency of a unified model while better preserving task-specific expertise through input-dependent routing at both the task and head levels. However, even the routing components of the state-of-the-art approach remain limited: the task routing is highly sensitive to the choice of routing layer, whereas the head routing relies on simple max-logit competition. To address these limitations, we propose \underline{m}ulti-layer \underline{a}ggregated rou\underline{t}ing \underline{e}vidence (MATE) as a unified mechanism for improving both task and head routing. Specifically, we first introduce a metric to quantify task routing separability across different layers. Empirical analysis of this metric reveals that different layers provide complementary evidence for identifying the tasks relevant to an input. Motivated by this observation, MATE selects multiple informative layers according to their routing separability and aggregates their layer-wise routing scores to produce a more robust signal that can be directly used for task selection. Moreover, we observe that this signal also encodes useful information about the suitability of the corresponding classification heads. This motivates us to further combine the MATE signal with the max-logit score to guide classification head selection. Extensive experiments on vision benchmarks demonstrate that MATE consistently achieves state-of-the-art performance.

\end{abstract}

\section{Introduction}

In recent years, fine-tuning large pretrained models has become
standard practice for downstream task specialization~\citep{devlin2019bert,neyshabur2020transfer,wortsman2022wiseft,ilharco2022patching}. As the number of tasks grows, however, maintaining a separate fine-tuned model for each task becomes increasingly costly~\citep{houlsby2019adapters,hu2022lora}. Conventional multi-task learning can reduce this burden by
jointly training a single model, but requires the original data
and incurs additional training costs~\citep{jin2022dataless}. By contrast, model merging offers an alternative route to consolidation by directly combining the parameters of existing fine-tuned models and reusing the task knowledge encoded in their weights~\citep{matena2022merging,ilharco2023taskarithmetic}. Its reduced dependence on training and the original datasets has attracted growing research interest in recent years~\citep{yang2024adamerging,zheng2025learningfrommodels,yang2026model}.

Early model merging research explored parameter averaging
to consolidate separately trained models~\citep{singh2020model,wortsman2022model,matena2022merging}.
In a seminal work, \citet{ilharco2023taskarithmetic}
introduced Task Arithmetic to compose task knowledge through
parameter updates rather than averaging complete model weights. It represents task-specific parameter changes relative to a shared pretrained model as task vectors and combines them to compose task capabilities. Building on this formulation, subsequent methods reduce interference among task vectors by removing redundant updates and resolving conflicts in the original parameter space~\citep{yadav2023ties,yu2024language}. More recent work extends this perspective to the subspace structure of task vectors, using singular value decomposition to identify singular directions and magnitudes. These efforts include exploiting low-rank structure~\citep{choi2024revisiting,gargiulo2025task,skorobogat2025subspace} and improving subspace alignment and spectral balance during merging~\citep{stoica2025model,marczak2025isotropic,panariello2026accurate,li2026esm,zhang2026dcmerge}.

Despite these advances, these methods share a static paradigm: a single merged model applies the same parameter combination to every input. This fixed combination can weaken task-specific directions and diminish the expertise needed for individual tasks. To better preserve task-specific expertise, dynamic model merging draws on the mixture-of-experts (MoE) idea, treating task vectors as experts to be selected and combined according to each input~\citep{yadav2024survey}. A straightforward solution retains complete task vectors and uses task identity to guide adaptive merging and prediction. While this preserves access to task-specific knowledge, its storage cost grows linearly with the number of tasks. More importantly, this solution relies on task identity at inference time, whereas the unknown-task scenario requires the model to infer task relevance and make predictions from the input alone. These considerations raise two challenges. First, how can task knowledge be retained at low cost? Second, how can input-dependent routing identify both the relevant task knowledge and the appropriate classification head without prior knowledge of the task identity?

WEMoE~\citep{tang2024wemoe} addresses the first challenge by retaining task-specific MLP weights for dynamic merging while statically merging the remaining parameters. However, its input-dependent merging weights rely on parametric routers that require training through test-time adaptation. Several subsequent studies further explore expert sparsification, shared and task-exclusive components, and selective dynamic merging across layers, but still rely on parametric routers that require additional training~\citep{lu2024twinmerging,ye2025mow,shen2025efficient}. 
To eliminate this training requirement, SMILE~\citep{tang2025zero} compresses task vectors into low-rank experts through singular value decomposition and introduces nonparametric routing based on projections of intermediate input representations onto task subspaces. However, it selects experts sequentially from lower to higher layers, allowing routing errors to propagate as later decisions depend on representations produced by earlier selections.
More importantly, WEMoE and SMILE address only part of the second challenge: they select and combine task knowledge according to the input, but their evaluations still rely on oracle classification heads specified by the true task identity. 

\begin{figure*}[t]
    \centering
    \begin{subfigure}[b]{0.44\textwidth}
        \centering
        \includegraphics[width=\linewidth]{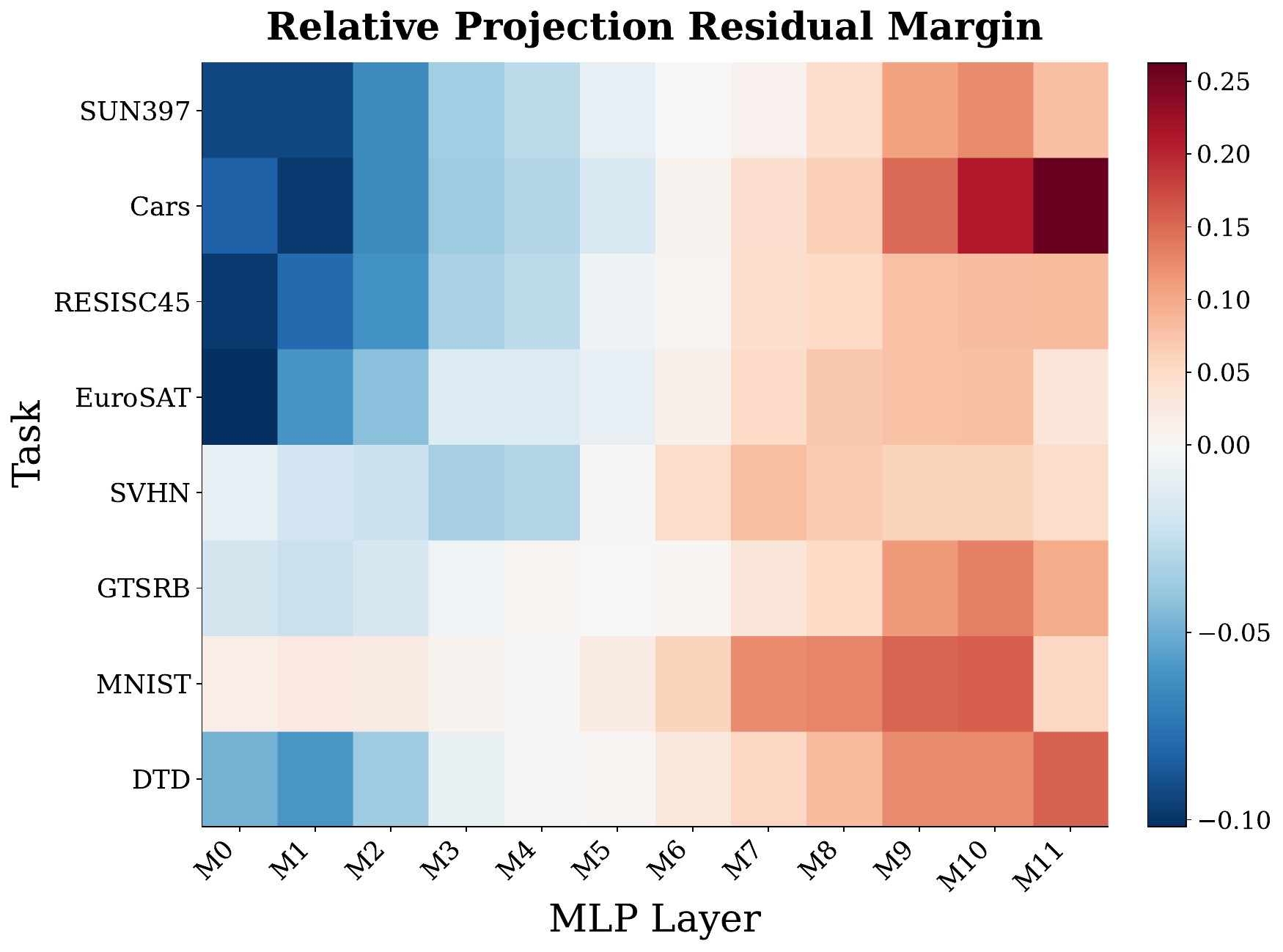}
        \caption{Layer-wise task routing separability}
        \label{fig:motivation-margin}
    \end{subfigure}
    \hfill
    \begin{subfigure}[b]{0.46\textwidth}
        \centering
        \includegraphics[width=\linewidth]{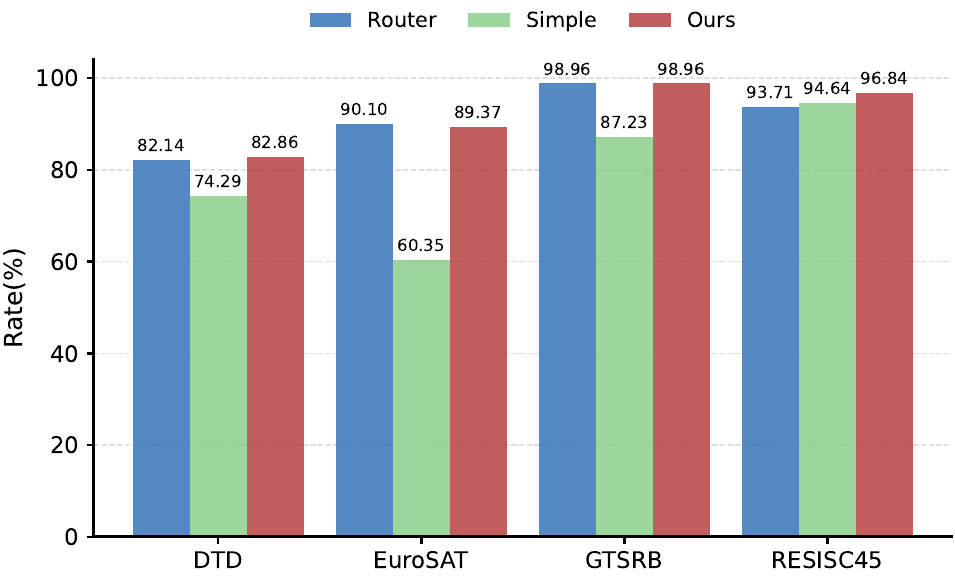}
        \caption{True-task selection rates among samples with multiple candidate tasks and the true task retained}
        \label{fig:motivation-head}
    \end{subfigure}
    \caption{
Empirical motivation for our method.
(a) Relative projection residual margins across MLP layers
of ViT-B/32 with 8 tasks.
(b) With the true task among multiple candidates, Router reports how often it has the highest routing score. Simple and Ours respectively report its head's win rate under simple and routing-guided max-logit competition.
More details are provided in Appendix~\ref{sec:head_selection_analysis}.
}
    \label{fig:motivation}
\end{figure*}

MASS~\citep{crisostomi2026mass} takes a different approach to task routing, decoupling task selection from adaptive merging through two-pass inference. It first constructs a fixed merged model using TSV-M~\citep{gargiulo2025task} and uses projection residuals at a single preselected layer to select candidate task subspaces, which are then merged for a second forward pass. Crucially, it extends routing to classification heads through max-logit competition, enabling prediction without prior knowledge of the task. MASS achieves state-of-the-art performance among existing methods.
However, it still has two key limitations in both task and head routing. Specifically, task routing is highly sensitive to layer selection, with different tasks favoring different layers. MASS reports cross-task accuracy variation of up to $40\%$ at some layers~\citep{crisostomi2026mass}. Meanwhile, head routing relies on simple max-logit competition, which can select an incorrect head even when the true task's subspace is retained among the candidates. For example, among EuroSAT samples with multiple candidate tasks including EuroSAT, simple max-logit competition selects the EuroSAT classification head in only $60.35\%$ of cases, as shown in Figure~\ref{fig:motivation-head}.

In this paper, we propose \underline{m}ulti-layer \underline{a}ggregated rou\underline{t}ing \underline{e}vidence (MATE) as a unified mechanism for improving both task and head routing. To improve task routing, we first investigate its sensitivity to layer selection, which calls for a quantitative measure of layer-wise task separability. The projection residuals used in MASS reflect input--task compatibility, but do not directly provide such a measure. Their magnitudes depend on representation scale, while a small residual for the true task does not ensure separation from competing tasks. We therefore normalize the residuals by the representation norm and measure the margin between the true task and its most competitive incorrect task. The resulting relative projection residual margin characterizes task separability across layers, with larger values indicating clearer separation. Based on this metric, our layer-wise analysis in Figure~\ref{fig:motivation-margin} reveals distinct separability patterns across tasks: layers that discriminate well for one task may be less informative for another. These patterns suggest that multiple layers provide complementary routing evidence. Motivated by this observation, MATE selects informative candidate layers according to their routing separability and aggregates their routing scores into a more robust signal for adaptive task subspace selection.

Moreover, since each classification head corresponds to a particular task, the routing signal identifying relevant task knowledge should also help distinguish suitable heads. We test this intuition on samples whose candidate sets contain multiple tasks, including the true task, as shown in Figure~\ref{fig:motivation-head}. Notably, on EuroSAT, the true task receives the highest routing score in $90.10\%$ of cases, whereas simple max-logit competition selects its head in only $60.35\%$. This gap shows that the true task's top routing rank does not imply that its classification head will win under simple max-logit competition. Interestingly, we further weight head logits by routing scores and find that this raises the correct-head selection rate to $89.37\%$, substantially narrowing the gap. These findings motivate us to incorporate the MATE signal into head routing, weighting head logits before max-logit selection so the same evidence supports both task subspace selection and classification head routing.

We evaluate MATE on visual model merging benchmarks comprising 8, 14, and 20 tasks using three CLIP~\citep{radford2021clip} backbones: ViT-B/32, ViT-B/16, and ViT-L/14. MATE consistently outperforms the evaluated baselines across all nine configurations, supporting the effectiveness of multi-layer routing aggregation and routing-guided head selection.

To summarize, our key contributions are as follows:
\begin{itemize}
    \item We introduce the relative projection residual margin to quantify layer-wise task routing separability, revealing task-dependent layer preferences and complementary routing evidence. We further show that routing scores provide useful information for classification head selection.
    \item We propose MATE, a unified mechanism for task and head routing in the unknown-task scenario. MATE selects informative layers and aggregates their routing scores for adaptive task subspace selection, then uses the same aggregated signal to guide classification head selection.
    \item We evaluate MATE across nine configurations spanning three CLIP ViT backbones and three task scales. MATE achieves the best model merging performance in all nine configurations, with maximum gains across the three task scales of 2.19, 1.84, and 0.69 percentage points over the strongest baseline on ViT-B/32, ViT-B/16, and ViT-L/14, respectively. Routing analyses and ablation studies further support the effectiveness of both designs.
\end{itemize}

\section{Preliminaries}
\label{sec:preliminaries}
In this section, we briefly review the basic ingredients underlying MATE: task vectors, TSV-M for subspace-based model merging, and the routing procedure used in MASS.

\paragraph{Task vectors.}
Let $\theta_{\mathrm{pre}}$ denote the pretrained model parameters and
$\{\theta_i\}_{i=1}^{T}$ the parameters of models independently fine-tuned
from $\theta_{\mathrm{pre}}$ on $T$ downstream tasks.
Following Task Arithmetic~\citep{ilharco2023taskarithmetic},
the parameter update for task $i$, referred to as its task vector, is
\begin{equation}
\Delta_i = \theta_i-\theta_{\mathrm{pre}}.
\label{eq:task_vector}
\end{equation}

\paragraph{TSV-M.}
Let $\Delta_i^{(\ell)}$ denote the matrix corresponding to task vector
$\Delta_i$ at layer $\ell$. TSV-M~\citep{gargiulo2025task} compresses
this matrix by retaining its leading singular components, yielding
$\Delta_i^{(\ell)} \approx
U_i^{(\ell)}\Sigma_i^{(\ell)}V_i^{(\ell)\top}$,
where $V_i^{(\ell)}$ contains the retained right singular vectors.
In MASS and our method, $V_i^{(\ell)}$ is used as the task subspace basis for routing.
At each layer, let $U$ and $V$ denote the concatenations of the retained left and right singular vectors across tasks, and let $\Sigma$ be the block-diagonal
matrix of their singular values. TSV-M orthogonalizes these bases
to reduce cross-task interference and reconstructs the merged update:
\begin{equation}
\Delta_{\mathrm{MT}}^{(\ell)}
=
\operatorname{orth}(U)\Sigma\operatorname{orth}(V)^\top,
\label{eq:tsv_merge}
\end{equation}
where $\operatorname{orth}(X)=PQ^\top$ for the reduced SVD
$X=PDQ^\top$.
The layer-wise updates form $\Delta_{\mathrm{MT}}$, yielding
$\theta_{\mathrm{MT}}=\theta_{\mathrm{pre}}+\alpha\Delta_{\mathrm{MT}}$,
where TSV-M sets $\alpha=1$.
Algorithm~\ref{alg:tsv_merge} summarizes the layer-wise merging procedure.
{
\setlength{\algoheightrule}{1.0pt}
\setlength{\algotitleheightrule}{0.5pt}
\IncMargin{1.5em}

\begin{algorithm}[t]
\RestyleAlgo{ruled}

\SetAlgoNoLine
\SetAlgoShortEnd
\LinesNumbered
\DontPrintSemicolon

\SetAlgoNlRelativeSize{-1}
\SetNlSty{normalfont}{}{:}
\SetNlSkip{0.4em}
\SetInd{0.5em}{1em}
\small

\SetKwInput{KwIn}{Input}
\SetKwInput{KwOut}{Output}

\caption{TSV-M~\citep{gargiulo2025task}}
\label{alg:tsv_merge}

\Indm
\KwIn{
Task vectors
$\{\Delta_i^{(\ell)}\}_{i=1}^{T}$ at layer $\ell$
}
\KwOut{Merged update $\Delta_{\mathrm{MT}}^{(\ell)}$}
\Indp

\For{$i=1,\ldots,T$}{
    Compute the reduced SVD
    $\Delta_i^{(\ell)}
    =U_i^{(\ell)}\Sigma_i^{(\ell)}V_i^{(\ell)\top}$\;

    Keep the leading $1/T$ fraction of singular components
    in $U_i^{(\ell)}$, $\Sigma_i^{(\ell)}$, and
    $V_i^{(\ell)}$\;
}

$U\leftarrow
[U_1^{(\ell)},\ldots,U_T^{(\ell)}],
\quad
V\leftarrow
[V_1^{(\ell)},\ldots,V_T^{(\ell)}]$\;

$\Sigma\leftarrow
\operatorname{block-diag}
(\Sigma_1^{(\ell)},\ldots,\Sigma_T^{(\ell)})$\;

Compute the SVDs
$U=P_U D_U Q_U^\top$ and
$V=P_V D_V Q_V^\top$\;

$U_{\perp}\leftarrow P_U Q_U^\top,
\quad
V_{\perp}\leftarrow P_V Q_V^\top$\;

$\Delta_{\mathrm{MT}}^{(\ell)}
\leftarrow U_{\perp}\Sigma V_{\perp}^{\top}$\;

\Return{$\Delta_{\mathrm{MT}}^{(\ell)}$}\;

\end{algorithm}
}
\paragraph{MASS.}
MASS~\citep{crisostomi2026mass} first constructs the fixed merged model
$\theta_{\mathrm{MT}}$ by setting
$\theta_{\mathrm{MT}}^{(\ell)}
=\theta_{\mathrm{pre}}^{(\ell)}
+\operatorname{TSV\text{-}M}
(\{\Delta_i^{(\ell)}\}_{i=1}^{T})$
at each layer, then passes the input $x$ through it.
Let $z_\ell(x)$ denote the intermediate representation entering
the weight matrix at layer $\ell$ of $\theta_{\mathrm{MT}}$.
MASS then uses $z_\ell(x)$ at a single preselected routing layer
to measure task relevance through the projection residual
\begin{equation}
r_{i,\ell}(x)
=
\left\|
z_\ell(x)
-
V_i^{(\ell)}V_i^{(\ell)\top}z_\ell(x)
\right\|_2.
\label{eq:projection_residual}
\end{equation}
\pdfdiff{
Smaller residuals indicate closer task matches.
Their standardized values
(Appendix~\ref{sec:routing_score_standardization})
yield a soft task distribution:
}
\begin{equation}
w_{\ell,i}(x)
=
\frac{
    \exp\!\left(-\pdfdiff{\widetilde{r}_{i,\ell}(x)}\right)
}{
    \sum_{j=1}^{T}
    \exp\!\left(-\pdfdiff{\widetilde{r}_{j,\ell}(x)}\right)
}.
\label{eq:layer_routing_score}
\end{equation}
MASS forms a candidate task set $\Omega(x)$ by thresholding these scores
with $\eta$ and retaining at most $K_T$ top-scoring tasks. It then
merges the selected task subspaces and performs a second forward pass
to obtain the final representation $z^{\mathrm{ada}}(x)$.
Let $h_i$ denote the classification head for task $i$, with class set
$\mathcal{C}_i$. MASS makes the final prediction through simple max-logit competition
among candidate classification heads.

\section{Method}
\label{sec:method}

In this section, we present multi-layer aggregated routing evidence (MATE) as a unified mechanism for task and head routing in the unknown-task scenario. We first quantify layer-wise task separability using the relative projection residual margin on a small set of proxy samples. We then select informative candidate routing layers and aggregate their routing scores into a task relevance signal for adaptive subspace selection. Finally, we use the same signal to guide the final decision among candidate classification heads.

\subsection{Layer-wise Routing Analysis}
\label{sec:overview}

The projection residual in Equation~(\ref{eq:projection_residual}) depends on both the alignment with a task subspace and the magnitude of the input representation. Differences in representation scale can therefore obscure comparisons across layers. To assess layer-wise task separability on a comparable scale, we first normalize the residual by the representation norm:
\begin{equation}
    \bar{r}_{i,\ell}(x)
    =
    \frac{
        r_{i,\ell}(x)
    }{
        \left\|z_\ell(x)\right\|_2+\epsilon
    }.
    \label{eq:relative_projection_residual}
\end{equation}
Here, $r_{i,\ell}(x)$ is the projection residual between the representation $z_\ell(x)$ and task $i$'s subspace at layer $\ell$, and $\epsilon$ is a constant for numerical stability.

For a sample $x$ from task $i$, the normalized residual of the true task alone still does not fully characterize the discriminative ability of a routing layer. Its separation from the most competitive incorrect task also matters. Therefore, we further define the sample-level relative projection residual margin as
\begin{equation}
    m_{i,\ell}(x)
    =
    \min_{j\neq i}
    \bar{r}_{j,\ell}(x)
    -
    \bar{r}_{i,\ell}(x).
    \label{eq:relative_residual_margin}
\end{equation}
When $m_{i,\ell}(x)>0$, the true task has the smallest relative projection residual. A larger margin indicates clearer separation between the true task and the most competitive incorrect task.
To characterize layer-wise task routing separability, we use a small proxy dataset $\mathcal{D}_i^{\mathrm{proxy}}$ for each task $i$ and compute its task-specific mean margin at layer $\ell$:
\begin{equation}
    A_{i,\ell}
    =
    \frac{1}{
        \left|\mathcal{D}_i^{\mathrm{proxy}}\right|
    }
    \sum_{x\in\mathcal{D}_i^{\mathrm{proxy}}}
    m_{i,\ell}(x).
    \label{eq:task_layer_margin}
\end{equation}

Figure~\ref{fig:motivation-margin} shows that layers with high routing separability for one task can be less informative for another. These task-dependent differences suggest complementary routing evidence across layers.
Additional layer-wise analyses across backbones and task scales are provided in Appendix~\ref{sec:additional_layer_analysis}.

\subsection{Candidate Layer Selection and Multi-Layer Routing Aggregation}
\label{sec:Multi_layer_routing}

Building on the preceding analysis, we now introduce MATE's
multi-layer routing aggregation mechanism.
Following MASS, we first construct $\theta_{\mathrm{MT}}$
using TSV-M~\citep{gargiulo2025task} for routing and cache
the compressed task subspaces for efficient reuse during
subsequent merging.
Rather than relying on a single layer, MATE selects multiple
informative layers according to their task separability and
aggregates their routing scores for candidate task selection.

\paragraph{Candidate layer selection.}
However, aggregating routing scores from all layers would also include evidence from layers with weak task separability. MATE therefore computes the task-averaged margin $\bar{A}_{\ell}=\frac{1}{T}\sum_{i=1}^{T}A_{i,\ell}$ for each available MLP routing layer $\ell\in\mathcal{L}$ and selects the top $K_L$ layers to form the candidate set:
\begin{equation}
\mathcal{L}_{\mathrm{cand}}
=
\operatorname{TopK}_{\ell\in\mathcal{L}}
\left(\bar{A}_{\ell}, K_L\right).
\label{eq:candidate_layer_selection}
\end{equation}

\paragraph{Multi-layer routing aggregation.}
Given $\mathcal{L}_{\mathrm{cand}}$, a single forward pass through $\theta_{\mathrm{MT}}$ obtains the representations $\{z_\ell(x)\}_{\ell\in\mathcal{L}_{\mathrm{cand}}}$, from which MATE computes the layer-wise routing scores using Equations~(\ref{eq:projection_residual}) and~(\ref{eq:layer_routing_score}). MATE then averages the routing scores over
$\mathcal{L}_{\mathrm{cand}}$ before candidate task selection,
retaining evidence that individual-layer thresholding or
Top-$K$ selection would discard:
\begin{equation}
    s_i(x)
    =
    \frac{1}{|\mathcal{L}_{\mathrm{cand}}|}
    \sum_{\ell\in\mathcal{L}_{\mathrm{cand}}}
    w_{\ell,i}(x).
    \label{eq:Multi_layer_fusion}
\end{equation}

The aggregated score $s_i(x)$ represents the relative support for task $i$ across the candidate layers. Based on this score, we perform task routing by constructing
the candidate task set $\Omega(x)$.
To be precise, we first retain tasks whose scores meet the threshold $\eta$,
forming $\Omega_{\eta}(x)=\{i\mid s_i(x)\geq\eta\}$.
Furthermore, we restrict this set to at most $K_T$ tasks with the
highest aggregated scores:
\begin{equation}
    \Omega(x)
    =
    \operatorname{TopK}
    \left(\Omega_{\eta}(x), s(x), K_T\right),
    \label{eq:routing_topk}
\end{equation}
where $s(x)=(s_1(x),\ldots,s_T(x))$.
If no task passes the threshold, i.e., $\Omega_{\eta}(x)=\varnothing$, we retain the task with the highest aggregated score by setting $\Omega(x)=\{\arg\max_i s_i(x)\}$.

After task routing, we merge the selected task subspaces to
construct the model for the current input. Note that this step is not intrinsically tied to our routing mechanism: once the candidate task set $\Omega(x)$ is determined, other compatible model merging methods could in principle be employed. To isolate the contribution of our routing design and ensure a fair comparison, we follow MASS in using TSV-M~\citep{gargiulo2025task} to merge the task subspaces associated with $\Omega(x)$ at each layer.

{
\setlength{\algoheightrule}{1.0pt}
\setlength{\algotitleheightrule}{0.5pt}
\IncMargin{1.5em}

\begin{algorithm}[t]
\RestyleAlgo{ruled}

\SetAlgoNoLine
\SetAlgoShortEnd
\LinesNumbered
\DontPrintSemicolon

\SetAlgoNlRelativeSize{-1}
\SetNlSty{normalfont}{}{:}
\SetNlSkip{0.4em}
\SetInd{0.5em}{1em}
\SetCommentSty{algcommentstyle}
\small

\SetKwInput{KwIn}{Input}
\SetKwInput{KwOut}{Output}

\caption{Dynamic Model Merging via MATE}
\label{alg:multi-layer-aggregated-routing-evidence}

\Indm
\KwIn{
sample $x$;
proxy sets $\{\mathcal{D}_i^{\mathrm{proxy}}\}_{i=1}^{T}$;
pretrained model $\theta_{\mathrm{pre}}$;
task vectors $\{\Delta_i\}_{i=1}^{T}$;
MLP routing layers $\mathcal{L}$ and candidate layer count $K_L$;
routing threshold $\eta$ and task limit $K_T$;
task heads $\{h_i\}_{i=1}^{T}$
}
\KwOut{Predicted class $c^\star$}
\Indp
Obtain the fixed merged model $\theta_{\mathrm{MT}}$ and
layer-wise subspace bases
$\{V_i^{(\ell)}\}_{i=1,\ldots,T;\,\ell\in\mathcal{L}}$
using TSV-M with $\theta_{\mathrm{pre}}$ and
$\{\Delta_i\}_{i=1}^{T}$
\tcp*[r]{Cache compressed task subspaces}

\tcp*[l]{\textbf{Offline: one-time candidate layer selection before inference}}

\For{$i=1,\ldots,T$}{
    \For{$x'\in\mathcal{D}_i^{\mathrm{proxy}}$}{
        $\{z_\ell(x')\}_{\ell\in\mathcal{L}}
        \leftarrow
        \mathrm{ForwardPass}
        (\theta_{\mathrm{MT}},x';\mathcal{L})$\;

        \For{$\ell\in\mathcal{L}$}{
            Compute $\bar{r}_{j,\ell}(x')$ for $j=1,\ldots,T$
            using Eq.~(\ref{eq:relative_projection_residual})\;

            Compute $m_{i,\ell}(x')$
            using Eq.~(\ref{eq:relative_residual_margin})\;
        }
    }

    Compute $A_{i,\ell}$ for $\ell\in\mathcal{L}$
    using Eq.~(\ref{eq:task_layer_margin})\;
}

$\bar{A}_{\ell}
\leftarrow
\frac{1}{T}\sum_{i=1}^{T}A_{i,\ell},
\quad \ell\in\mathcal{L}$\;

Select $\mathcal{L}_{\mathrm{cand}}$
using Eq.~(\ref{eq:candidate_layer_selection})\;

\BlankLine
\tcp*[l]{\textbf{Online: multi-layer routing and prediction}}

$\{z_\ell(x)\}_{\ell\in\mathcal{L}_{\mathrm{cand}}}
\leftarrow
\mathrm{ForwardPass}
(\theta_{\mathrm{MT}},x;\mathcal{L}_{\mathrm{cand}})$\;

\For{$\ell\in\mathcal{L}_{\mathrm{cand}}$}{
    Compute $r_{i,\ell}(x)$ for $i=1,\ldots,T$
    using Eq.~(\ref{eq:projection_residual})\;

    Compute $w_{\ell,i}(x)$ for $i=1,\ldots,T$
    using Eq.~(\ref{eq:layer_routing_score})\;
}

$s_i(x)
\leftarrow
\frac{1}{|\mathcal{L}_{\mathrm{cand}}|}
\sum_{\ell\in\mathcal{L}_{\mathrm{cand}}}
w_{\ell,i}(x),
\quad i=1,\ldots,T$
\tcp*[r]{Aggregate routing scores}

$\Omega_{\eta}(x)
\leftarrow
\{i\mid s_i(x)\geq\eta\}$\;

$\Omega(x)
\leftarrow
\operatorname{TopK}
\bigl(\Omega_{\eta}(x),s(x),K_T\bigr)$\;

\lIf{$\Omega_{\eta}(x)=\varnothing$}{
    $\Omega(x)\leftarrow\{\arg\max_i s_i(x)\}$
}

$\theta_{\mathrm{adp}}
\leftarrow
\theta_{\mathrm{pre}}
+
\operatorname{TSV\text{-}M}
\bigl(\{\Delta_i\}_{i\in\Omega(x)}\bigr)$
\tcp*[r]{Apply to each layer using cached subspaces}

$z^{\mathrm{ada}}(x)
\leftarrow
\mathrm{ForwardPass}(\theta_{\mathrm{adp}},x)$\;

$\tilde{o}_i(x)
\leftarrow
s_i(x)\,h_i\bigl(z^{\mathrm{ada}}(x)\bigr),
\quad i\in\Omega(x)$
\tcp*[r]{Routing-guided head decision}

Select $(i^\star,c^\star)$
using Eq.~(\ref{eq:final_prediction})\;

\Return{$c^\star$}\;

\end{algorithm}
}

\subsection{Routing-Guided Head Decision}
\label{sec:guided_head}

Having selected and merged the candidate task subspaces, we now
turn to head routing to determine which candidate classification
head should make the final prediction.
The second forward pass through the dynamic model yields
$z^{\mathrm{ada}}(x)$, from which each candidate head produces
logits $o_i(x)=h_i(z^{\mathrm{ada}}(x))$ for $i\in\Omega(x)$.
A straightforward choice is to use simple max-logit competition
following MASS.

However, Figure~\ref{fig:motivation-head} shows that this rule
can select an incorrect head even when the true task is retained,
while routing scores remain informative for head selection.
Motivated by these findings, we retain the aggregated task
relevance scores after subspace selection and use them to guide
head competition by weighting each candidate head's logits:
\begin{equation}
    \tilde{o}_i(x)=s_i(x)o_i(x),
    \qquad i\in\Omega(x).
    \label{eq:guided_head}
\end{equation}

Finally, we select the task $i^\star$ and class $c^\star$ corresponding to the largest weighted logit across all candidate heads:
\begin{equation}
    (i^\star,c^\star)
    =
    \operatorname*{arg\,max}_{i\in\Omega(x),\,c\in\mathcal{C}_i}
    \tilde{o}_i(x)[c].
    \label{eq:final_prediction}
\end{equation}

Algorithm~\ref{alg:multi-layer-aggregated-routing-evidence}
summarizes the complete procedure of dynamic model merging via MATE.

\section{Experiments}
\label{sec:experiments}

In this section, we evaluate MATE against static and dynamic model merging baselines across three ViT backbones and three task scales. We further investigate its routing behavior and examine the
contributions of multi-layer aggregation and routing-guided
head decisions through routing analyses, ablation studies,
and \pdfdiff{candidate layer selection analyses}.

\subsection{Experimental Setup}

Following the experimental setup of MASS~\citep{crisostomi2026mass}, we evaluate model merging in the 8-task, 14-task, and 20-task visual classification settings using CLIP~\citep{radford2021clip} with ViT-B/32, ViT-B/16, and ViT-L/14 backbones. More details on the datasets for each task are provided in Appendix~\ref{sec:benchmarks_datasets}. We report the average test accuracy across tasks. Baselines include static merging methods: Task Arithmetic (TA)~\citep{ilharco2023taskarithmetic}, TSV-M~\citep{gargiulo2025task}, Iso-C and Iso-CTS~\citep{marczak2025isotropic}, and dynamic merging methods: WEMoE~\citep{tang2024wemoe}, SMILE~\citep{tang2025zero}, and MASS~\citep{crisostomi2026mass}. To evaluate WEMoE and SMILE without task identity, we also follow MASS in selecting classification heads by majority voting over their layer-wise routing decisions. Independently fine-tuned models serve as a reference. All merging methods use the same task models and evaluation configurations.

For MATE, we construct each proxy dataset $\mathcal{D}_i^{\mathrm{proxy}}$ by randomly sampling 32 examples from task $i$'s training set. We select $K_L=5$ MLP routing layers with the highest relative projection residual margins averaged across tasks. Following MASS, we set the routing threshold to $\eta=0.20$ and the maximum number of candidate tasks to $K_T=3$ in all main experiments. Appendix~\ref{sec:hyperparameter_sensitivity} examines the sensitivity of MATE to these hyperparameters.

\subsection{Multi-Task Model Merging Performance}

\begin{table}[t]
    \centering
    \setlength{\tabcolsep}{3.8pt}
    \renewcommand{\arraystretch}{1.08}

    \caption{
    Average classification accuracy (\%) of model merging methods across three ViT backbones and different task scales.
    Bold indicates the best result among model merging methods.
    }
    \label{tab:main_results}

    \resizebox{\linewidth}{!}{%
    \begin{tabular}{clccccccccc}
        \toprule[1.0pt]

        &
        \multirow{2}{*}{\textbf{Method}}
        & \multicolumn{3}{c}{\textbf{ViT-B/32}}
        & \multicolumn{3}{c}{\textbf{ViT-B/16}}
        & \multicolumn{3}{c}{\textbf{ViT-L/14}} \\

        \cmidrule(lr){3-5}
        \cmidrule(lr){6-8}
        \cmidrule(lr){9-11}

        &
        & \textbf{8 tasks}
        & \textbf{14 tasks}
        & \textbf{20 tasks}
        & \textbf{8 tasks}
        & \textbf{14 tasks}
        & \textbf{20 tasks}
        & \textbf{8 tasks}
        & \textbf{14 tasks}
        & \textbf{20 tasks} \\

        \midrule[0.7pt]

        & Fine-tuned
        & 90.31 & 89.07 & 89.51
        & 92.42 & 91.37 & 91.92
        & 94.23 & 93.43 & 94.00 \\

        \mainTableDash

        \multirow{4}{*}{\rotatebox{90}{\textbf{Static}}}
        & Task Arithmetic
        & 68.82 & 64.62 & 64.03
        & 72.96 & 70.64 & 69.01
        & 84.39 & 80.37 & 76.85 \\

        &
        TSV-M
        & 83.21 & 78.62 & 75.55
        & 85.52 & 81.38 & 78.80
        & 91.25 & 88.78 & 87.49 \\

        &
        Iso-C
        & 82.85 & 78.44 & 73.23
        & 87.47 & 79.82 & 75.29
        & 92.58 & 89.64 & 86.82 \\

        &
        Iso-CTS
        & 82.03 & 80.58 & 76.99
        & 88.67 & 84.11 & 80.69
        & 92.81 & 91.08 & 89.24 \\

        \mainTableDash

        \multirow{4}{*}{\rotatebox[origin=c]{90}{\textbf{Dynamic}}}
        & WEMoE
        & 88.87 & 74.09 & 68.61
        & 90.72 & 76.26 & 66.12
        & 92.92 & 71.34 & 68.53 \\

        &
        SMILE
        & 83.27 & 77.29 & 72.87
        & 87.89 & 81.78 & 79.51
        & 91.31 & 86.61 & 84.93 \\

        &
        MASS
        & 87.08 & 82.90 & 81.17
        & 90.60 & 87.83 & 81.11
        & 92.97 & 90.95 & 90.84 \\

        &
        \ourscell{\textbf{MATE}}
        & \ourscell{\textbf{88.91}}
        & \ourscell{\textbf{85.09}}
        & \ourscell{\textbf{82.25}}
        & \ourscell{\textbf{91.44}}
        & \ourscell{\textbf{88.55}}
        & \ourscell{\textbf{82.95}}
        & \ourscell{\textbf{93.66}}
        & \ourscell{\textbf{91.58}}
        & \ourscell{\textbf{91.00}} \\

        \bottomrule[1.0pt]
    \end{tabular}%
    }
\end{table}

Table~\ref{tab:main_results} reports average classification accuracy across three ViT backbones in the 8-task, 14-task, and 20-task settings. MATE achieves the highest accuracy among model merging methods in all nine configurations, outperforming both static and dynamic baselines. Across the three task scales, its maximum gains over the strongest baseline are 2.19, 1.84, and 0.69 percentage points on ViT-B/32, ViT-B/16, and ViT-L/14, respectively. As the number of tasks increases, input-dependent merging requires selecting relevant knowledge from a broader set of task models. MATE retains its performance advantage in the 14-task and 20-task settings across all three backbones, supporting its effectiveness across the evaluated task scales.

\subsection{Routing Performance}
\label{sec:routing_performance}

To better understand MATE's performance gains, we also compare its
multi-layer routing aggregation with MASS's single-layer routing
using top-1 task accuracy and task survival.
Top-1 task accuracy measures the proportion of samples for which the highest-scoring task matches the input's true task. Task survival measures the proportion of samples for which the true task is retained in the candidate set $\Omega(x)$, allowing its knowledge to participate in subsequent merging even when it is not ranked first.

\begin{table}[t]
    \centering
    \setlength{\tabcolsep}{4.2pt}
    \renewcommand{\arraystretch}{1.08}

    \caption{
    Routing performance of the single-layer baseline and MATE across ViT backbones and task scales.
    \textbf{(a)} Top-1 task accuracy;
    \textbf{(b)} Task survival.
    Bold indicates the better result.
    }
    \label{tab:routing_results}

    \textbf{(a) Top-1 Task Accuracy (\%)} \\[3pt]

    \resizebox{\linewidth}{!}{%
    \begin{tabular}{lccccccccc}
        \toprule[1.0pt]

        \multirow{2}{*}{\textbf{Method}}
        & \multicolumn{3}{c}{\textbf{ViT-B/32}}
        & \multicolumn{3}{c}{\textbf{ViT-B/16}}
        & \multicolumn{3}{c}{\textbf{ViT-L/14}} \\

        \cmidrule(lr){2-4}
        \cmidrule(lr){5-7}
        \cmidrule(lr){8-10}

        & \textbf{8 tasks}
        & \textbf{14 tasks}
        & \textbf{20 tasks}
        & \textbf{8 tasks}
        & \textbf{14 tasks}
        & \textbf{20 tasks}
        & \textbf{8 tasks}
        & \textbf{14 tasks}
        & \textbf{20 tasks} \\

        \midrule[0.7pt]

        Single-Layer (MASS)
        & 97.34 & 94.25 & 84.08
        & 99.11 & 96.86 & 83.08
        & 99.46 & 97.99 & 90.17 \\

        \rowcolor{oursblue}
        \textbf{MATE}
        & \textbf{98.69}
        & \textbf{96.27}
        & \textbf{86.28}
        & \textbf{99.32}
        & \textbf{97.67}
        & \textbf{84.50}
        & \textbf{99.63}
        & \textbf{98.32}
        & \textbf{91.12} \\

        \bottomrule[1.0pt]
    \end{tabular}%
    }

    \vspace{8pt}

    \textbf{(b) Task Survival (\%)} \\[3pt]

    \resizebox{\linewidth}{!}{%
    \begin{tabular}{lccccccccc}
        \toprule[1.0pt]

        \multirow{2}{*}{\textbf{Method}}
        & \multicolumn{3}{c}{\textbf{ViT-B/32}}
        & \multicolumn{3}{c}{\textbf{ViT-B/16}}
        & \multicolumn{3}{c}{\textbf{ViT-L/14}} \\

        \cmidrule(lr){2-4}
        \cmidrule(lr){5-7}
        \cmidrule(lr){8-10}

        & \textbf{8 tasks}
        & \textbf{14 tasks}
        & \textbf{20 tasks}
        & \textbf{8 tasks}
        & \textbf{14 tasks}
        & \textbf{20 tasks}
        & \textbf{8 tasks}
        & \textbf{14 tasks}
        & \textbf{20 tasks} \\

        \midrule[0.7pt]

        Single-Layer (MASS)
        & 99.67 & 96.72 & 91.57
        & 99.88 & 98.65 & 89.15
        & 99.91 & 99.24 & 95.19 \\

        \rowcolor{oursblue}
        \textbf{MATE}
        & \textbf{99.73}
        & \textbf{98.00}
        & \textbf{92.66}
        & \textbf{99.93}
        & \textbf{98.90}
        & \textbf{91.14}
        & \textbf{99.95}
        & \textbf{99.47}
        & \textbf{96.14} \\

        \bottomrule[1.0pt]
    \end{tabular}%
    }
\end{table}

Table~\ref{tab:routing_results} shows that MATE outperforms MASS in both top-1 task accuracy and task survival across all nine configurations. On ViT-B/32, top-1 task accuracy reaches 98.69\%, 96.27\%, and 86.28\% in the 8-task, 14-task, and 20-task settings, respectively, with a gain of 2.20 percentage points in the 20-task setting. The improvements extend to ViT-B/16 and ViT-L/14. Task survival also improves consistently, rising from 89.15\% to 91.14\% on ViT-B/16 in the 20-task setting.

As the number of tasks increases, top-1 task accuracy declines more sharply than task survival. For example, in the ViT-B/16 20-task setting, MATE achieves a top-1 task accuracy of 84.50\% and a task survival of 91.14\%. This gap shows that the true task can remain in the candidate set $\Omega(x)$ without receiving the highest routing score. Once retained, its task subspace participates in adaptive merging and its classification head enters the final prediction competition, preserving the opportunity for a correct prediction despite a task-ranking error.

Together, these results indicate that multi-layer aggregation
provides a more reliable signal for task selection than
single-layer routing across all nine configurations.
By incorporating task-relevant evidence that a single layer
can overlook, it reduces reliance on any individual layer.
The resulting improvements in task identification and true-task
retention help explain MATE's gains in final classification accuracy.

\subsection{Routing Separability of Attention and MLP Layers}
\label{sec:layer_type_separability}

\begin{wrapfigure}{r}{0.43\textwidth}
    \centering
    \includegraphics[width=\linewidth]{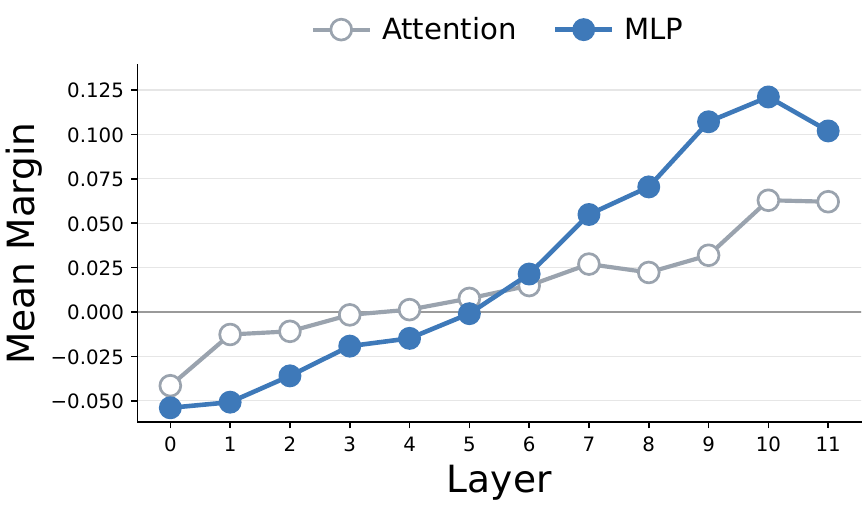}
    \caption{
    Task-averaged relative projection residual margins
    in the ViT-B/32 8-task setting.
    }
    \label{fig:task_layer_margin}
\end{wrapfigure}

Our main experiments use MLP layers for routing. To examine this choice, we additionally compare attention and MLP layers in the ViT-B/32 8-task setting using the relative projection residual margin averaged across tasks at each Transformer block. 
Figure~\ref{fig:task_layer_margin} shows that attention layers have higher mean margins in the early blocks, but these remain negative or near zero, indicating limited task separability. In blocks 6--11, MLP layers instead achieve larger positive margins than attention layers within the same blocks. Since each MLP sublayer receives hidden states already updated by self-attention, it also provides a natural location for routing within the block. Motivated by these observations, we use MLP layers as the candidate pool and select informative layers, drawing routing evidence from distinct blocks while retaining a single representation source per block.

\subsection{Ablation Study}
\label{sec:ablation}

\begin{table}[t]
    \centering
    \setlength{\tabcolsep}{3.6pt}
    \renewcommand{\arraystretch}{1.10}

    \caption{
        Average classification accuracy (\%) with different combinations
        of multi-layer routing aggregation and routing-guided head decision.
        $\checkmark$ and $\times$ indicate enabled and disabled components. Bold indicates the best result in each configuration.
    }
    \label{tab:ablation}

    \resizebox{\linewidth}{!}{%
    \begin{tabular}{ccccccccccc}
        \toprule[1.0pt]

        \multirow{2}{*}{%
            \shortstack[c]{%
                \textbf{Multi-Layer Routing}\\
                \textbf{Aggregation}%
            }%
        }
        & \multirow{2}{*}{%
            \shortstack[c]{%
                \textbf{Routing-Guided Head}\\
                \textbf{Decision}%
            }%
        }
        & \multicolumn{3}{c}{\textbf{ViT-B/32}}
        & \multicolumn{3}{c}{\textbf{ViT-B/16}}
        & \multicolumn{3}{c}{\textbf{ViT-L/14}} \\

        \cmidrule(lr){3-5}
        \cmidrule(lr){6-8}
        \cmidrule(lr){9-11}

        &
        & \textbf{8 tasks}
        & \textbf{14 tasks}
        & \textbf{20 tasks}
        & \textbf{8 tasks}
        & \textbf{14 tasks}
        & \textbf{20 tasks}
        & \textbf{8 tasks}
        & \textbf{14 tasks}
        & \textbf{20 tasks} \\

        \midrule[0.7pt]

        $\times$
        & $\times$
        & 87.08 & 82.90 & 81.17
        & 90.60 & 87.83 & 81.11
        & 92.97 & 90.95 & 90.84 \\

        $\times$
        & $\checkmark$
        & 88.52 & 83.77 & 81.43
        & 91.31 & 88.17 & 80.95
        & 93.55 & 91.39 & 90.55 \\

        $\checkmark$
        & $\times$
        & 87.34 & 84.34 & 81.82
        & 90.35 & 88.20 & \textbf{83.36}
        & 93.00 & 91.08 & \textbf{91.11} \\

        \rowcolor{oursblue}
        $\checkmark$
        & $\checkmark$
        & \textbf{88.91}
        & \textbf{85.09}
        & \textbf{82.25}
        & \textbf{91.44}
        & \textbf{88.55}
        & 82.95
        & \textbf{93.66}
        & \textbf{91.58}
        & 91.00 \\

        \bottomrule[1.0pt]
    \end{tabular}%
    }
\end{table}

We evaluate four combinations of multi-layer routing aggregation and routing-guided head decision. Disabling aggregation retains single-layer routing, while disabling head guidance uses simple max-logit competition among candidate heads. Disabling both recovers MASS, while enabling both yields the full MATE. As shown in Table~\ref{tab:ablation}, enabling multi-layer routing aggregation alone improves classification accuracy in eight of the nine configurations. In the ViT-B/16 20-task setting, accuracy rises from 81.11\% to 83.36\%, a gain of 2.25 percentage points. Complementing the direct evaluation of top-1 task accuracy and task survival in Section~\ref{sec:routing_performance}, this ablation shows that multi-layer aggregation also benefits final classification accuracy.

Enabling routing-guided head decision alone improves performance in seven of the nine configurations. For example, accuracy reaches 88.52\% and 93.55\% in the ViT-B/32 and ViT-L/14 8-task settings. These gains indicate that routing scores retain useful task relevance information after candidate selection and can guide the final competition among classification heads. Nevertheless, small decreases of 0.16 and 0.29 percentage points occur in the ViT-B/16 and ViT-L/14 20-task settings. Compared with the 14-task setting, this larger task set introduces additional visual overlap, with CIFAR10 added alongside CIFAR100 and STL10, and EMNIST and KMNIST alongside MNIST. Such overlap can reduce the discriminative power of routing scores within the candidate set. When task rankings become less reliable, routing-score weighting may shift the balance among candidate heads and slightly reduce accuracy.

With both components enabled, MATE outperforms MASS in all nine configurations and achieves the highest accuracy among the ablation variants in seven, indicating complementary benefits in most configurations. In the ViT-B/16 and ViT-L/14 20-task settings, however, adding routing-guided head decision to multi-layer routing aggregation slightly reduces accuracy. This is consistent with the preceding analysis: less discriminative routing scores among visually related candidate tasks can limit the benefit of head weighting, even after multi-layer aggregation.

\subsection{\pdfdiff{Candidate Layer Selection Analysis}}
\label{sec:layer_selection_comparison}

\begin{table}[!t]
\centering
\small
\setlength{\tabcolsep}{10pt}
\renewcommand{\arraystretch}{1.08}

\caption{\pdfdiff{
Stability of candidate layer selection under proxy resampling.
Each proxy size is evaluated across nine configurations
with five sampling seeds per configuration.
A match indicates the same candidate layer set as in the
corresponding main experiment, regardless of internal ordering.
}}
\label{tab:proxy_sampling_stability}

\pdfdiff{
\begin{tabular}{@{}ccc@{}}
    \toprule
    \textbf{Proxy examples per task}
    & \textbf{Matching runs}
    & \textbf{Match rate (\%)} \\
    \midrule
    8
    & 44/45
    & 97.8 \\
    16
    & 45/45
    & 100.0 \\
    32 (default)
    & 45/45
    & 100.0 \\
    \bottomrule
\end{tabular}
}
\end{table}

\paragraph{\pdfdiff{Proxy sampling stability.}}
\pdfdiff{
We repeat candidate layer selection using 8, 16, and 32
proxy examples per task, with five sampling seeds for each
size across all nine configurations.
Table~\ref{tab:proxy_sampling_stability} shows that 134 of
the 135 runs recover the candidate layer set used in the
main experiments.
All 45 runs at each of the 16- and 32-example settings
recover the original set.
The only mismatch occurs with eight examples and changes
a single selected layer.
These results support stable layer selection with small
proxy sets.
Since proxy data are used only for offline layer selection,
identical layer sets preserve the subsequent inference
configuration when task models and other conditions are fixed.
\par
}

\paragraph{\pdfdiff{Routing strategy comparison.}}
\pdfdiff{
We next compare five routing strategies, keeping the task
models, TSV-M merging, and routing-guided head decision fixed.
Single-Layer (MASS) uses MASS's original routing layer.
Proxy-Best Single and MATE Top-5 use the same proxy samples
to select the one and five layers with the highest
task-averaged margins $\bar{A}_{\ell}$, respectively.
Random-5 averages routing scores from five randomly selected
MLP layers, while All-MLP uses all MLP layers.
\par
}

\begin{table}[t]
\centering
\small
\setlength{\tabcolsep}{4pt}
\renewcommand{\arraystretch}{1.08}

\caption{\pdfdiff{
Average classification accuracy (\%) under different routing
layer selection and aggregation strategies.
All variants use routing-guided head decision.
Random-5 reports mean $\pm$ standard deviation over three
layer-selection seeds.
Bold indicates the highest reported accuracy in each configuration.
}}
\label{tab:layer_selection_comparison}

\pdfdiff{
\begin{tabular}{@{}lcccccc@{}}
    \toprule
\makecell[c]{\textbf{Backbone}}
& \makecell[c]{\textbf{Tasks}}
& \makecell[c]{\textbf{Single-Layer}\\\textbf{(MASS)}}
& \makecell[c]{\textbf{Proxy-Best}\\\textbf{Single}}
& \makecell[c]{\textbf{Random-5}}
& \makecell[c]{\textbf{All-MLP}}
& \makecell[c]{\textbf{MATE}\\\textbf{Top-5}} \\
\midrule

    \multirow{3}{*}{ViT-B/32}
    & 8
    & 88.52 & 88.20 & $86.86 \pm 2.45$ & 87.22
    & \textbf{88.91} \\
    & 14
    & 83.77 & 83.65 & $83.40 \pm 1.70$ & 84.55
    & \textbf{85.09} \\
    & 20
    & 81.43 & 77.87 & $78.59 \pm 1.04$ & 80.89
    & \textbf{82.25} \\
    \midrule

    \multirow{3}{*}{ViT-B/16}
    & 8
    & 91.31 & 91.31 & $90.78 \pm 0.50$ & 91.05
    & \textbf{91.44} \\
    & 14
    & 88.17 & 83.97 & $86.14 \pm 3.16$ & 88.54
    & \textbf{88.55} \\
    & 20
    & 80.95 & 80.51 & $79.99 \pm 3.05$ & 81.36
    & \textbf{82.95} \\
    \midrule

    \multirow{3}{*}{ViT-L/14}
    & 8
    & 93.55 & 93.72 & $93.77 \pm 0.05$
    & \textbf{93.80} & 93.66 \\
    & 14
    & 91.39 & 91.66 & $92.21 \pm 0.40$
    & \textbf{92.45} & 91.58 \\
    & 20
    & 90.55 & 90.00 & $\mathbf{91.29} \pm 0.98$
    & 90.53 & 91.00 \\

    \bottomrule
\end{tabular}
}
\end{table}

\pdfdiff{
Table~\ref{tab:layer_selection_comparison} shows that MATE Top-5
outperforms Single-Layer (MASS) in all nine configurations and
Proxy-Best Single in seven, supporting the benefit of aggregation
beyond proxy-based single-layer selection. It also outperforms Random-5 and All-MLP in all six ViT-B configurations, but not consistently on ViT-L/14.
The relative projection residual margin thus supports layer
selection, although individual-layer separability does not
guarantee the most accurate layer combination.
\par
}

\pdfdiff{
Appendix~\ref{sec:aggregation_order} further compares aggregation
before and after softmax.
Inference efficiency and resource costs are reported in
Appendix~\ref{sec:efficiency_costs}.
\par
}

\section{Conclusion and Future Work}
\label{sec:conclusion}

In this paper, we present MATE, a unified and improved mechanism for task and head routing in dynamic model merging without prior task information. Our analysis reveals that different layers provide complementary evidence for task identification and that routing scores also inform classification head selection. We first introduce the relative projection residual margin to quantify layer-wise task separability. Based on this measure, MATE selects informative layers and aggregates their routing scores into a task relevance signal for adaptive task subspace selection and routing-guided head decisions. Experiments across three CLIP ViT backbones and three task scales show that MATE achieves the best model merging accuracy among the evaluated methods in all nine configurations, alongside consistent improvements in task identification and true-task retention.

There are still some directions for future work. First, MATE relies on a small proxy dataset for layer selection. It is interesting to explore fully data-free methods for
estimating layer-wise separability. Second, our ablation study shows that the gains from head weighting vary across task configurations. Thus, exploring alternative or complementary evidence for head routing may lead to more consistent improvements in final performance.





\bibliography{iclr2027_conference}
\bibliographystyle{iclr2027_conference}

\clearpage
\appendix

\section{Metric Definitions for Routing and Head Selection}
\label{sec:head_selection_analysis}

We conduct the analysis in Figure~\ref{fig:motivation-head}
using ViT-B/32 in the 8-task setting.
For each task $i$, all three metrics are computed over
$\mathcal{S}_i=\{x\in\mathcal{D}_i^{\mathrm{test}}
\mid |\Omega(x)|>1,\ i\in\Omega(x)\}$,
where $\mathcal{D}_i^{\mathrm{test}}$ denotes its test set.

\paragraph{Router.}
Given routing scores $s_j(x)$, Router reports the proportion
of samples for which the true task ranks highest:
\begin{equation}
R_i^{\mathrm{Router}}
=
\frac{1}{|\mathcal{S}_i|}
\sum_{x\in\mathcal{S}_i}
\mathbf{1}\!\left[
i=\operatorname*{arg\,max}_{j\in\Omega(x)}s_j(x)
\right].
\label{eq:head_analysis_router}
\end{equation}

\paragraph{Simple and Ours.}
Let $o_j(x)[c]$ denote the logit for class
$c\in\mathcal{C}_j$ from head $j$.
Simple and Ours respectively select heads through simple and
routing-guided max-logit competition:
\begin{align}
\hat{i}_{\mathrm{Simple}}(x)
&=
\operatorname*{arg\,max}_{j\in\Omega(x)}
\max_{c\in\mathcal{C}_j}o_j(x)[c],
\label{eq:head_analysis_simple}\\
\hat{i}_{\mathrm{Ours}}(x)
&=
\operatorname*{arg\,max}_{j\in\Omega(x)}
\max_{c\in\mathcal{C}_j}s_j(x)o_j(x)[c].
\label{eq:head_analysis_ours}
\end{align}
For both, we report the true-task head's win rate:
\begin{equation}
R_i^{a}
=
\frac{1}{|\mathcal{S}_i|}
\sum_{x\in\mathcal{S}_i}
\mathbf{1}\!\left[\hat{i}_{a}(x)=i\right],
\qquad
a\in\{\mathrm{Simple},\mathrm{Ours}\}.
\label{eq:head_analysis_rates}
\end{equation}

\section{Additional Layer-wise Routing Analysis}
\label{sec:additional_layer_analysis}

\begin{figure}[htbp]
    \centering

    \begin{subfigure}[t]{0.82\linewidth}
        \centering
        \includegraphics[width=\linewidth]
        {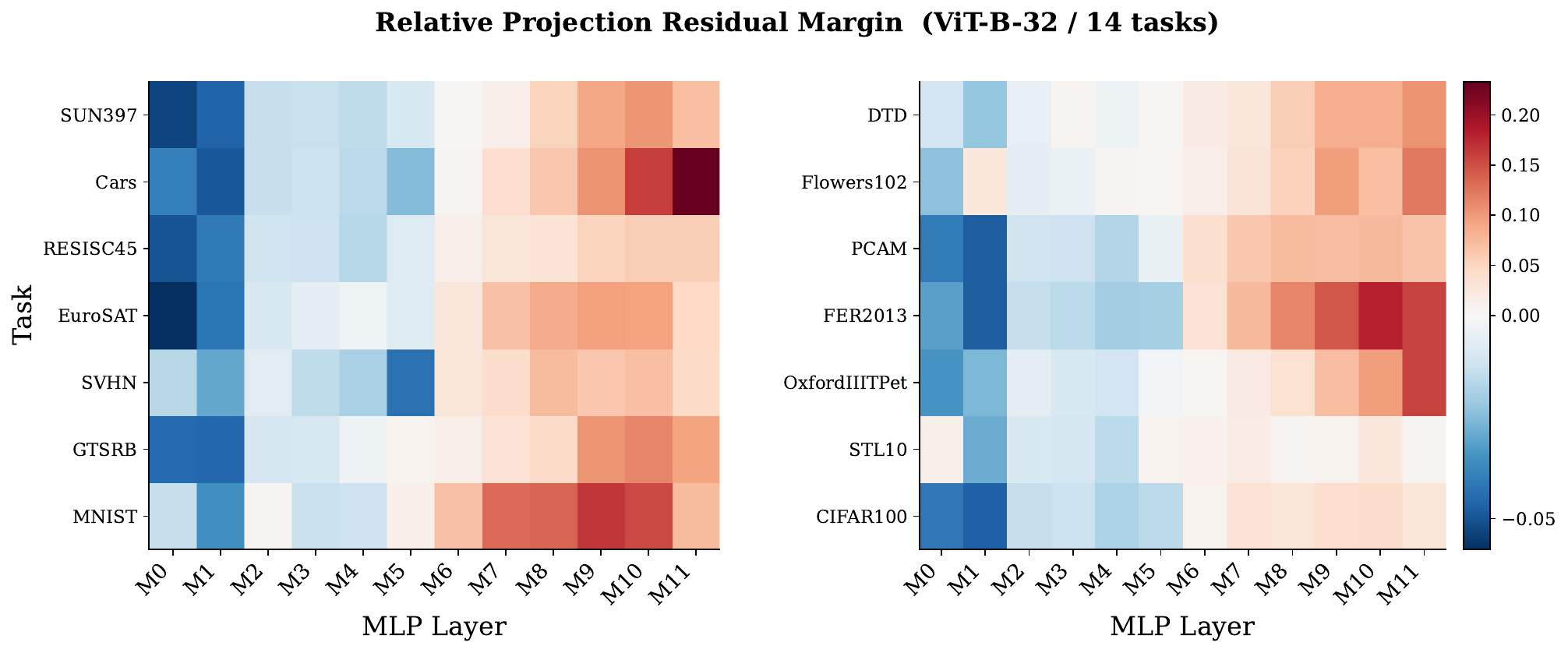}
        \caption{14-task}
        \label{fig:margin_b32_14tasks}
    \end{subfigure}

    \par\medskip

    \begin{subfigure}[t]{0.82\linewidth}
        \centering
        \includegraphics[width=\linewidth]
        {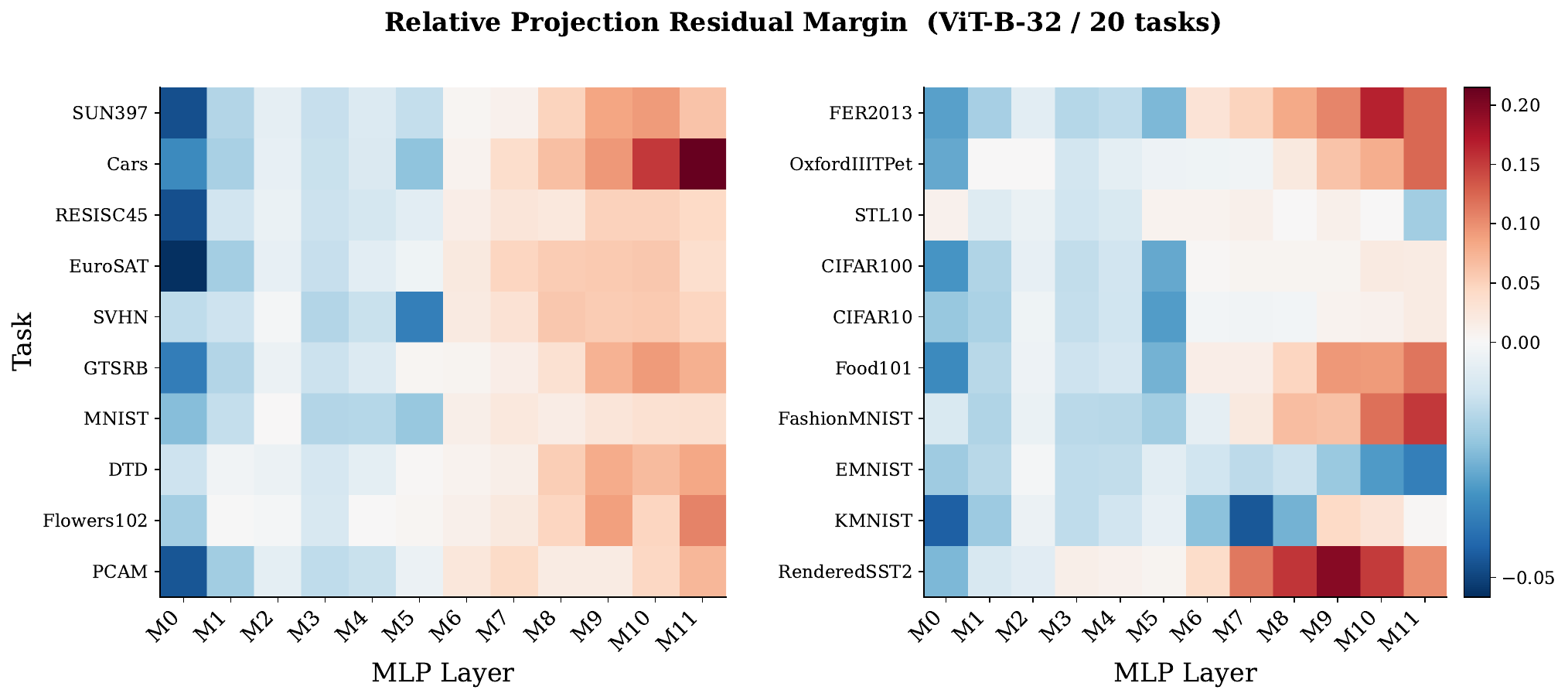}
        \caption{20-task}
        \label{fig:margin_b32_20tasks}
    \end{subfigure}

    \caption{
        Relative projection residual margins across
        MLP layers for ViT-B/32.
    }
    \label{fig:margin_b32_additional}
\end{figure}

\begin{figure}[htbp]
    \centering

    \begin{subfigure}[t]{0.49\linewidth}
        \centering
        \includegraphics[width=\linewidth]
        {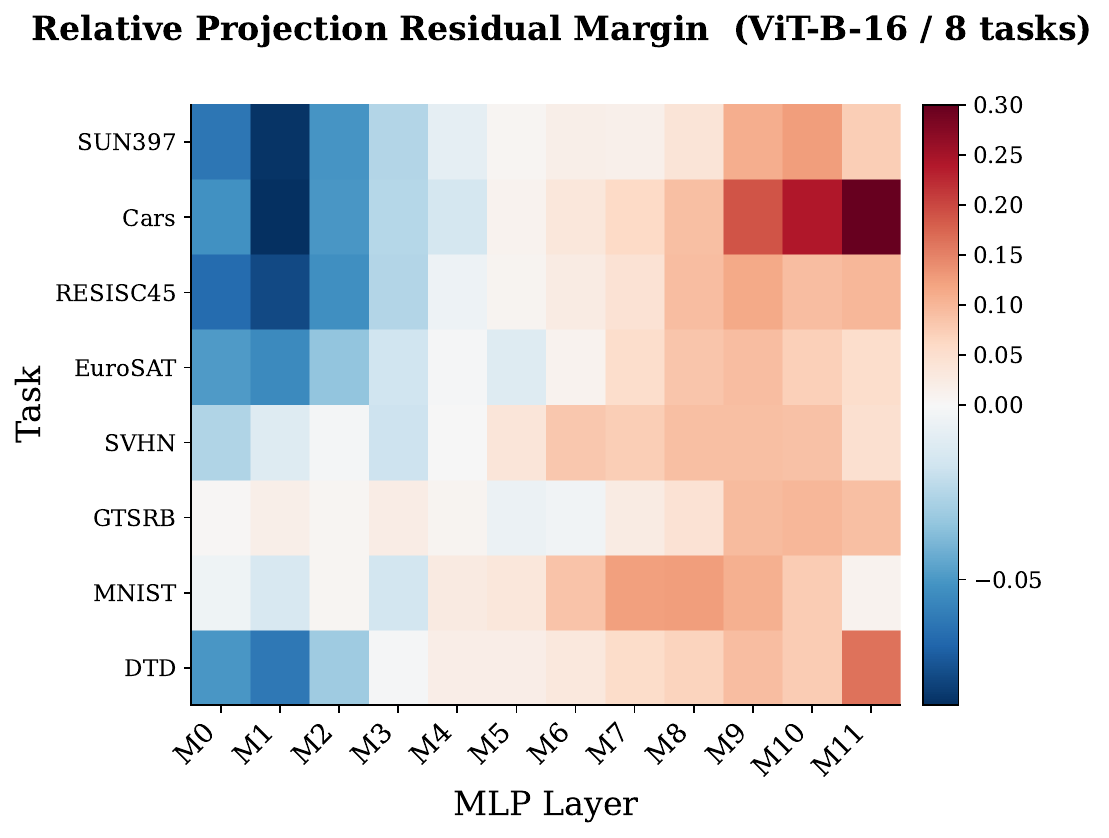}
        \caption{8-task}
        \label{fig:margin_b16_8tasks}
    \end{subfigure}

    \par\smallskip

    \begin{subfigure}[t]{0.81\linewidth}
        \centering
        \includegraphics[width=\linewidth]
        {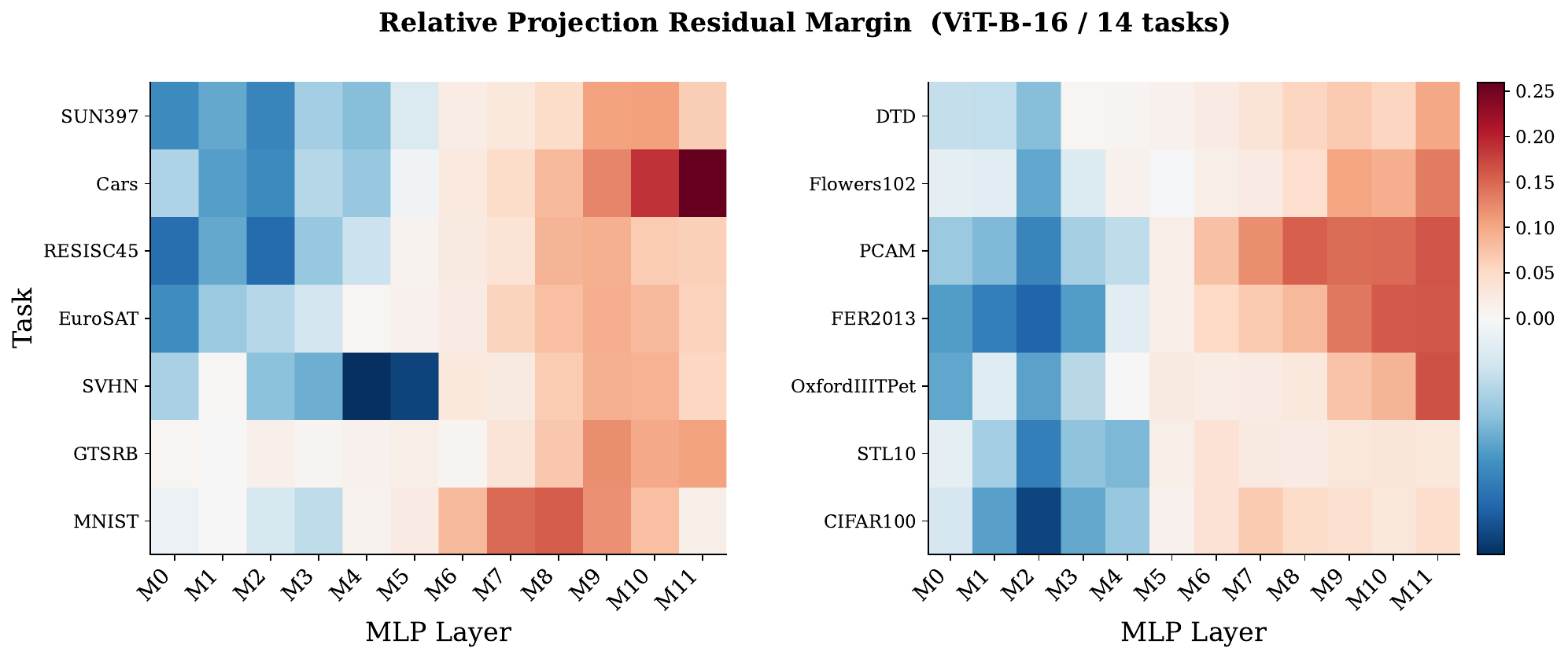}
        \caption{14-task}
        \label{fig:margin_b16_14tasks}
    \end{subfigure}

    \par\smallskip

    \begin{subfigure}[t]{0.81\linewidth}
        \centering
        \includegraphics[width=\linewidth]
        {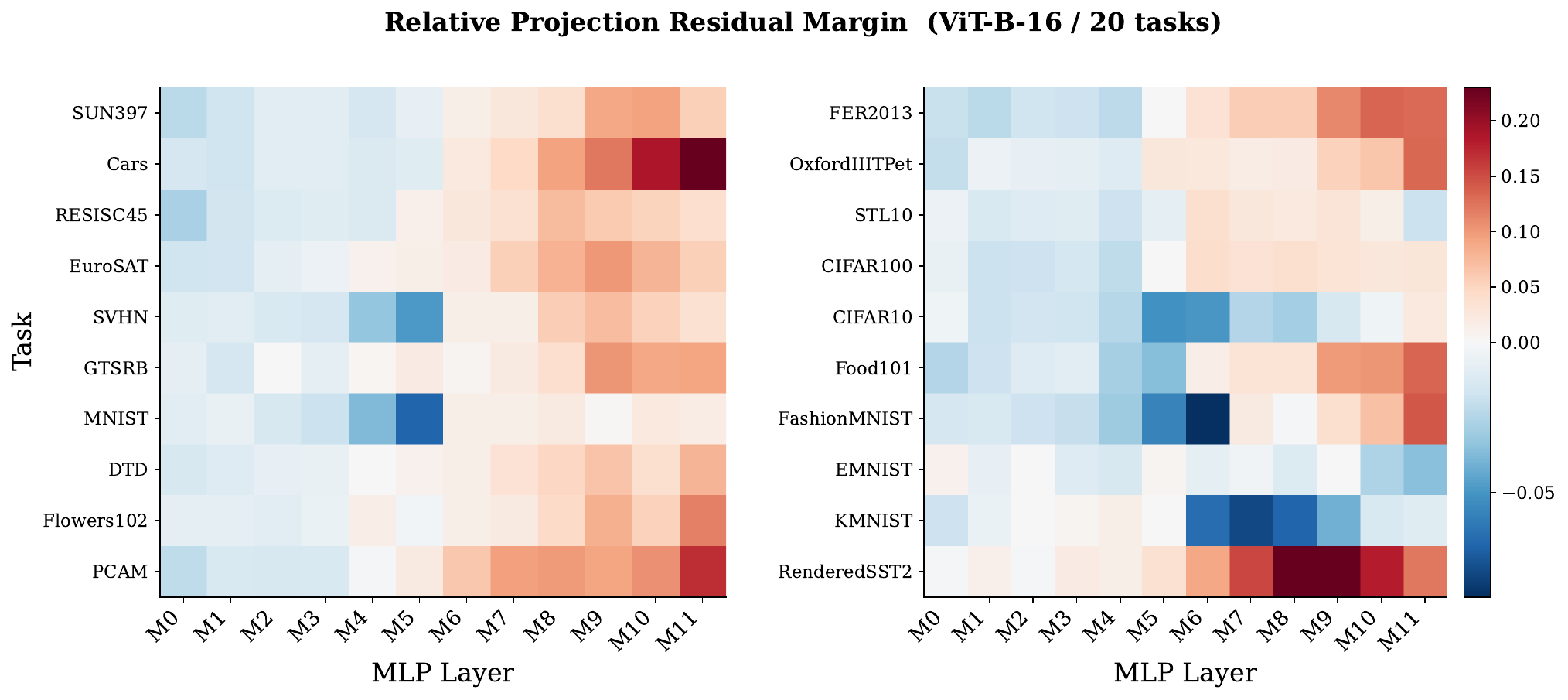}
        \caption{20-task}
        \label{fig:margin_b16_20tasks}
    \end{subfigure}

    \caption{
        Relative projection residual margins across
        MLP layers for ViT-B/16.
    }
    \label{fig:margin_b16_additional}
\end{figure}

\begin{figure}[htbp]
    \centering
    \includegraphics[width=0.66\linewidth]
    {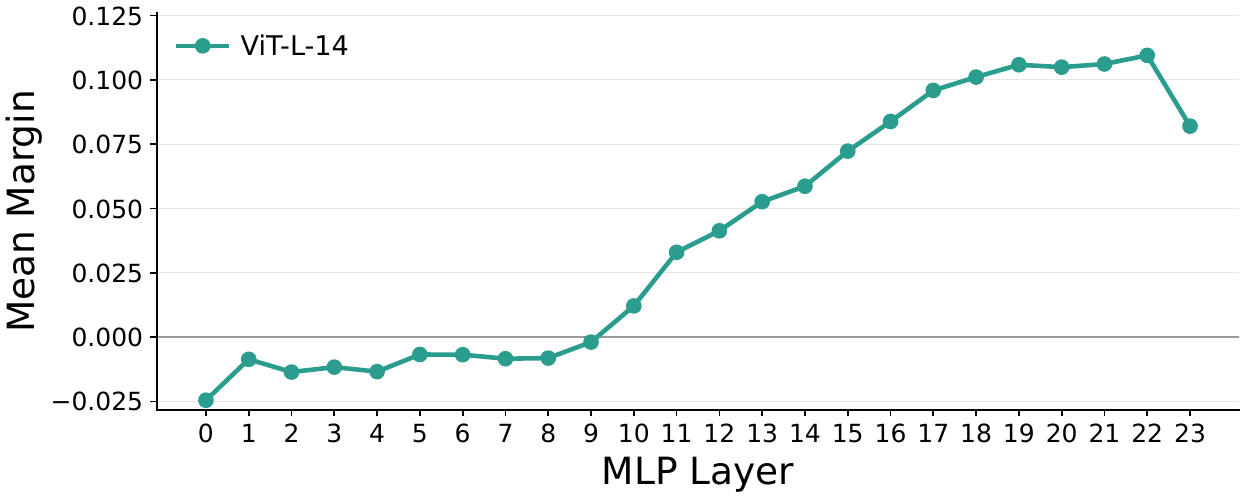}
    \caption{
        Relative projection residual margins averaged across
        tasks at each MLP layer of ViT-L/14 in the 20-task setting.
    }
    \label{fig:margin_vitl14_20tasks}
\end{figure}

We extend the analysis in Figure~\ref{fig:motivation-margin}
to additional task scales and backbones.
Figures~\ref{fig:margin_b32_additional}
and~\ref{fig:margin_b16_additional} show heatmaps for ViT-B/32
with 14 and 20 tasks and ViT-B/16 with 8, 14, and 20 tasks. 
Each cell reports $A_{i,\ell}$ from
Equation~(\ref{eq:task_layer_margin}), with larger values
indicating clearer task separation on average.
The heatmaps further illustrate task-dependent layer preferences.
For example, in both 20-task settings, Cars exhibits large positive
margins in the final MLP layers, whereas EMNIST has lower margins
in these layers. These contrasting patterns indicate that
layers informative for one task may be less useful for another,
further motivating the aggregation of routing evidence from
multiple informative layers.

Additionally, we examine ViT-L/14 in the 20-task setting.
For a compact view across its 24 MLP layers,
Figure~\ref{fig:margin_vitl14_20tasks} plots the task-averaged
margin $\bar{A}_{\ell}$, which is negative in early layers
and generally increases in later layers.

\section{Benchmarks and Datasets}
\label{sec:benchmarks_datasets}

\begin{table}[htbp]
    \centering
    \small
    \renewcommand{\arraystretch}{1.35}
    \setlength{\tabcolsep}{6pt}

    \caption{
        Datasets used in the three model merging benchmarks.
        The final column lists the task scales at which each dataset is included.
    }
    \label{tab:benchmark_datasets}

    \begin{tabularx}{0.70\linewidth}{
        @{}
        >{\raggedright\arraybackslash}X
        c
        @{}
    }
        \toprule
        \textbf{Dataset} & \textbf{Task scales} \\
        \midrule

        Cars~\citep{krause2013cars}
        & \multirow{8}{*}{8, 14, 20} \\
        DTD~\citep{cimpoi2014dtd} & \\
        EuroSAT~\citep{helber2019eurosat} & \\
        GTSRB~\citep{stallkamp2011gtsrb} & \\
        MNIST~\citep{lecun1998mnist} & \\
        RESISC45~\citep{cheng2017resisc45} & \\
        SUN397~\citep{xiao2016sun} & \\
        SVHN~\citep{netzer2011svhn} & \\

        \midrule

        CIFAR100~\citep{krizhevsky2009cifar}
        & \multirow{6}{*}{14, 20} \\
        STL10~\citep{coates2011stl10} & \\
        Flowers102~\citep{nilsback2008flowers} & \\
        OxfordIIITPet~\citep{parkhi2012pets} & \\
        PCAM~\citep{veeling2018pcam} & \\
        FER2013~\citep{goodfellow2013fer} & \\

        \midrule

        CIFAR10~\citep{krizhevsky2009cifar}
        & \multirow{6}{*}{20} \\
        Food101~\citep{bossard2014food101} & \\
        FashionMNIST~\citep{xiao2017fashionmnist} & \\
        EMNIST~\citep{cohen2017emnist} & \\
        KMNIST~\citep{clanuwat2018kmnist} & \\
        RenderedSST2~\citep{socher2013sst} & \\

        \bottomrule
    \end{tabularx}
\end{table}

We evaluate MATE on three visual classification benchmarks containing 8, 14, and 20 tasks, following MASS~\citep{crisostomi2026mass}. The 8-task benchmark follows the task collection used in Task Arithmetic~\citep{ilharco2023taskarithmetic}. Each larger benchmark retains all tasks from the preceding setting and adds six datasets. Table~\ref{tab:benchmark_datasets} lists all 20 datasets and their benchmark membership. The benchmarks cover natural images, remote sensing imagery, handwritten symbols, and medical images. Alongside this diversity, the larger task collections include related recognition problems, such as object recognition across CIFAR10, CIFAR100, and STL10, and handwritten symbol recognition across MNIST, EMNIST, and KMNIST. These settings therefore evaluate model merging across both visually distinct and related tasks.

\section{Hyperparameter Sensitivity}
\label{sec:hyperparameter_sensitivity}

\subsection{\pdfdiff{Routing Threshold $\eta$ and Maximum Candidate Task Count $K_T$}}
\label{sec:task_routing_hyperparameters}

\begin{figure}[t]
    \centering
    \begin{subfigure}[t]{0.43\linewidth}
        \centering
        \includegraphics[width=\linewidth]
        {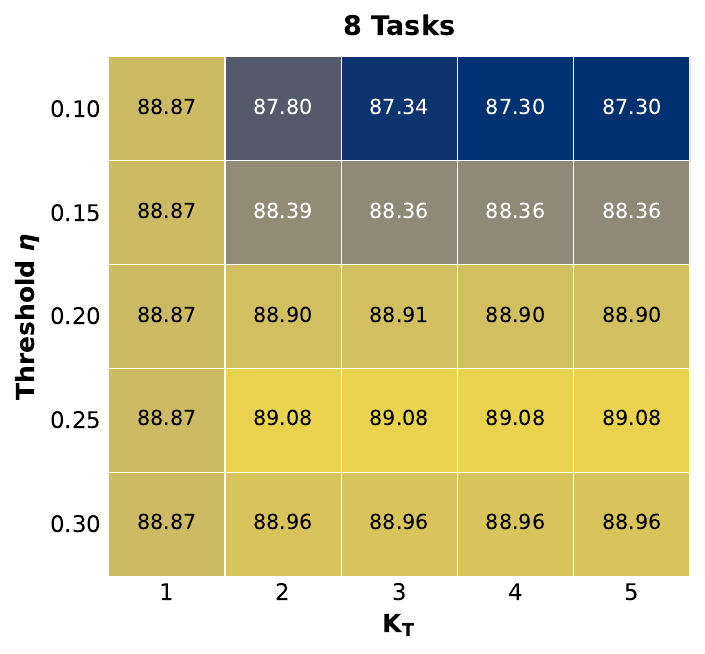}
        \caption{8-task}
        \label{fig:hyperparameter_sensitivity_8task}
    \end{subfigure}
    \hspace{0.03\linewidth}
    \begin{subfigure}[t]{0.43\linewidth}
        \centering
        \includegraphics[width=\linewidth]
        {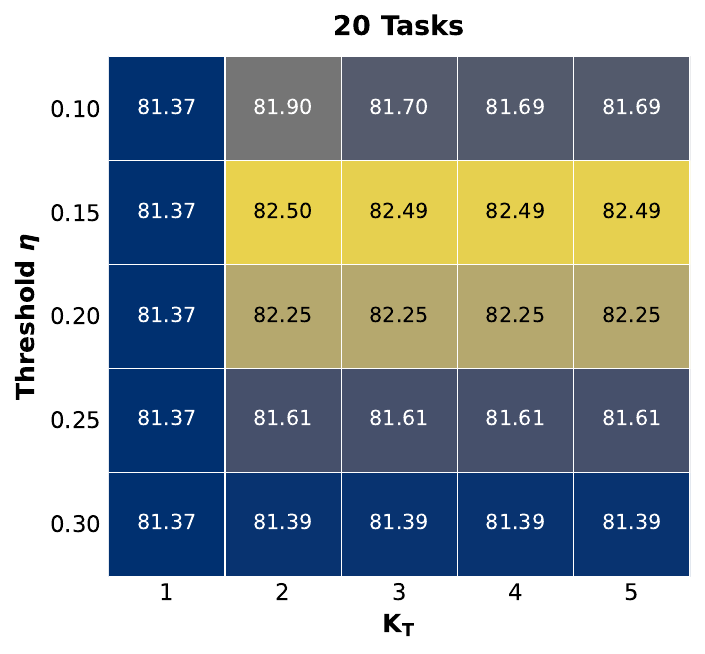}
        \caption{20-task}
        \label{fig:hyperparameter_sensitivity_20task}
    \end{subfigure}

    \caption{
    Sensitivity to the routing threshold $\eta$ and the maximum
    candidate task count $K_T$ on ViT-B/32.
    }
    \label{fig:hyperparameter_sensitivity}
\end{figure}

We further analyze the effects of the routing threshold $\eta$
and the maximum number of candidate tasks $K_T$ in the ViT-B/32
8-task and 20-task settings.
As shown in Figure~\ref{fig:hyperparameter_sensitivity}, MATE
maintains stable performance across a broad parameter range,
with favorable results near $\eta=0.25$ and $\eta=0.15$ in the
two settings, indicating that the preferred routing threshold
varies with task scale.

$K_T$ controls the maximum size of the candidate task set
$\Omega(x)$.
When the additional tasks have routing scores below $\eta$,
increasing $K_T$ does not enlarge the actual candidate set,
so the subsequent adaptive merging and predictions remain unchanged.
For example, for all $K_T\in\{2,3,4,5\}$, the 8-task setting
achieves $89.08\%$ at $\eta=0.25$, while the 20-task setting
achieves $82.25\%$ at $\eta=0.20$.

Our default setting, $\eta=0.20$ and $K_T=3$, lies in a
high-performing region for both task scales, with accuracy
only 0.17 and 0.25 percentage points below the best results
in the 8-task and 20-task settings.

\subsection{\pdfdiff{Candidate Layer Count $K_L$}}
\label{sec:candidate_layer_count}

\begin{table}[!htbp]
\centering
\small
\setlength{\tabcolsep}{4pt}
\renewcommand{\arraystretch}{1.08}

\caption{\pdfdiff{
Average classification accuracy (\%) for different candidate
layer counts $K_L$.
All denotes aggregation over all MLP layers.
}}
\label{tab:candidate_layer_count}

\pdfdiff{
\begin{tabular}{@{}lcccccccc@{}}
    \toprule
    \multirow{2}{*}{\textbf{Backbone}}
    & \multirow{2}{*}{\textbf{Tasks}}
    & \multicolumn{7}{c}{\textbf{Candidate layer count} $K_L$} \\
    \cmidrule(lr){3-9}
    &
    & \textbf{1}
    & \textbf{2}
    & \textbf{3}
    & \textbf{4}
    & \textbf{5}
    & \textbf{8}
    & \textbf{All} \\
    \midrule

    \multirow{3}{*}{ViT-B/32}
    & 8
    & 88.20 & 88.50 & 88.27 & 88.75
    & \textbf{88.91} & 88.87 & 87.22 \\
    & 14
    & 83.65 & 83.36 & 84.06 & 84.55
    & 85.09 & \textbf{85.46} & 84.55 \\
    & 20
    & 77.87 & 80.79 & 81.86 & 81.98
    & \textbf{82.25} & 80.99 & 80.89 \\
    \midrule

    \multirow{3}{*}{ViT-B/16}
    & 8
    & 91.31 & 91.25 & 91.00 & 91.31
    & \textbf{91.44} & 91.29 & 91.05 \\
    & 14
    & 83.97 & 87.53 & 87.51 & 88.13
    & 88.55 & \textbf{89.01} & 88.54 \\
    & 20
    & 80.51 & \textbf{83.50} & 83.21 & 83.08
    & 82.95 & 81.77 & 81.36 \\
    \midrule

    \multirow{3}{*}{ViT-L/14}
    & 8
    & 93.72 & 93.65 & 93.71 & 93.68
    & 93.66 & \textbf{93.82} & 93.80 \\
    & 14
    & 91.66 & 91.56 & 91.42 & 91.35
    & 91.58 & 92.02 & \textbf{92.45} \\
    & 20
    & 90.00 & 90.41 & 90.74 & 90.77
    & 91.00 & \textbf{91.41} & 90.53 \\
    \midrule

    \textbf{Average}
    & --
    & 86.77 & 87.84 & 87.98 & 88.18
    & \textbf{88.38} & 88.29 & 87.82 \\

    \bottomrule
\end{tabular}
}
\end{table}

\pdfdiff{
We examine the effect of the candidate layer count $K_L$
across all nine configurations.
Using the layer rankings obtained from the default proxy sets,
we vary $K_L$ while keeping the task models, candidate task
selection rule, TSV-M merging, and routing-guided head decision
unchanged.
For each $K_L$, routing scores are averaged over the layers
with the highest task-averaged margins $\bar{A}_{\ell}$.
\par
}

\pdfdiff{
Table~\ref{tab:candidate_layer_count} shows that increasing
$K_L$ from one to five raises the average accuracy across
configurations from 86.77\% to 88.38\%, a gain of 1.61
percentage points.
Further increasing the count to eight or all MLP layers
yields 88.29\% and 87.82\%, respectively.
Thus, aggregating more layers does not necessarily improve
classification accuracy.

The best layer count varies across configurations.
Our default $K_L=5$ achieves the highest average accuracy
among the tested choices while using fewer routing layers
than the eight-layer and All-MLP alternatives.
These results support using five candidate layers as a
shared default across backbones and task scales.
\par
}

\pdfdiff{
\section{Routing Score Standardization}
\label{sec:routing_score_standardization}

Following the \texttt{topk} routing mode in the public MASS
implementation~\citep{crisostomi2026mass}, we apply Z-score
standardization before computing the layer-wise routing scores.
For each input $x$ at layer $\ell$, let
$\widetilde{r}_{i,\ell}(x)$ denote the standardized projection
residual for task $i$:
\begin{equation*}
    \widetilde{r}_{i,\ell}(x)
    =
    \frac{
        r_{i,\ell}(x)-\mu_{\ell}(x)
    }{
        \sigma_{\ell}(x)+\epsilon
    },
\end{equation*}
where $\mu_{\ell}(x)$ and $\sigma_{\ell}(x)$ are the mean and
standard deviation of the residuals across all $T$ tasks
for the current input at layer $\ell$, and $\epsilon$ ensures
numerical stability.

The standardized residuals are converted into a soft task
distribution using Equation~\eqref{eq:layer_routing_score}.
MATE then uniformly averages these distributions across
candidate layers using Equation~\eqref{eq:Multi_layer_fusion}.
Thus, aggregation operates on task probabilities derived
from standardized residuals rather than on raw residuals.
\par
}

\section{\pdfdiff{Aggregation Before and After Softmax}}
\label{sec:aggregation_order}

\begin{table}[!htbp]
\centering
\small
\setlength{\tabcolsep}{6pt}
\renewcommand{\arraystretch}{1.08}

\caption{\pdfdiff{
Average classification accuracy (\%) under different routing
aggregation strategies.
The final row reports the equally weighted mean across
the nine configurations.
Bold indicates the highest accuracy in each row.
}}
\label{tab:aggregation_order}

\pdfdiff{
\begin{tabular}{@{}lcccc@{}}
    \toprule
    \makecell[c]{\textbf{Backbone}}
    & \makecell[c]{\textbf{Tasks}}
    & \makecell[c]{\textbf{Post-Softmax}\\\textbf{(MATE)}}
    & \makecell[c]{\textbf{Pre-Softmax}\\\textbf{Standardized}}
    & \makecell[c]{\textbf{Pre-Softmax}\\\textbf{Raw}} \\
    \midrule

    \multirow{3}{*}{ViT-B/32}
    & 8
    & 88.91 & 88.91 & \textbf{88.96} \\
    & 14
    & 85.09 & \textbf{85.16} & 84.83 \\
    & 20
    & \textbf{82.25} & 82.24 & 82.11 \\
    \midrule

    \multirow{3}{*}{ViT-B/16}
    & 8
    & \textbf{91.44} & 91.42 & 91.30 \\
    & 14
    & 88.55 & \textbf{88.74} & 88.47 \\
    & 20
    & 82.95 & 83.12 & \textbf{83.20} \\
    \midrule

    \multirow{3}{*}{ViT-L/14}
    & 8
    & 93.66 & 93.66 & \textbf{93.72} \\
    & 14
    & 91.58 & 91.61 & \textbf{91.63} \\
    & 20
    & 91.00 & \textbf{91.01} & 90.71 \\
    \midrule

    \textbf{Average}
    & --
    & 88.38 & \textbf{88.43} & 88.33 \\
    \bottomrule
\end{tabular}
}
\end{table}

\pdfdiff{
We examine whether routing aggregation benefits depend on
the placement of softmax.
All variants use the same Top-5 candidate layers, task models,
candidate task selection rule, TSV-M merging, and
routing-guided head decision.

Post-Softmax follows MATE by applying Z-score standardization
and softmax independently at each layer, then averaging the
resulting task probabilities.
Pre-Softmax Standardized applies Z-score standardization
within each layer, averages the negative standardized
residuals across candidate layers, and then applies softmax.
Pre-Softmax Raw averages the raw negative projection
residuals before applying softmax.
\par
}

\pdfdiff{
Table~\ref{tab:aggregation_order} shows that the three
strategies achieve similar average accuracies of 88.38\%,
88.43\%, and 88.33\%, differing by at most 0.10 percentage points.
These results suggest that multi-layer routing benefits do not depend on softmax aggregation order in the tested configurations. 
Together with Table~\ref{tab:layer_selection_comparison},
they support MATE’s contribution of selecting and combining task-discriminative evidence from multiple layers, rather than establishing a universally superior aggregation operator. 
\par
}

\section{\pdfdiff{Inference Efficiency and Resource Costs}}
\label{sec:efficiency_costs}
\pdfdiff{
Compared with MASS, MATE introduces one-time offline
proxy-based layer selection and online multi-layer routing
aggregation.
Since layer selection is performed only once, we focus
our efficiency evaluation on online end-to-end inference
and deployment resource costs.
The task-vector SVD preprocessing shared by both methods
is also performed offline and excluded from the reported
inference latency.
\par
}

\pdfdiff{
We compare MASS and MATE on an NVIDIA RTX A6000 using FP32
precision and batch size 1.
For each configuration, we sampled 240 test images evenly across tasks. MASS and MATE processed the same images in the same order, with measurements repeated three times.
Both methods use MASS's original dynamic task-group caching mechanism, which stores merged models indexed by candidate task sets. Cache hits reuse the stored models, while misses trigger TSV-M merging. The cache capacity and clearing policy remain identical.

Table~\ref{tab:efficiency_costs} reports online inference
and deployment resource costs.
End-to-end latency includes both backbone forward passes,
routing, dynamic merging, and classification head selection.
Disk storage includes model weights, classification heads,
and SVD/TSV representations, excluding runtime caches.
\par
}

\begin{table}[!htbp]
\centering
\small
\setlength{\tabcolsep}{4pt}
\renewcommand{\arraystretch}{1.08}

\caption{\pdfdiff{
Online inference and deployment resource costs.
}}
\label{tab:efficiency_costs}

\pdfdiff{
\begin{tabular}{@{}llcccc@{}}
    \toprule
    \makecell[c]{\textbf{Configuration}}
    & \makecell[c]{\textbf{Method}}
    & \makecell[c]{\textbf{E2E}\\\textbf{(ms)}}
    & \makecell[c]{\textbf{Routing}\\\textbf{(ms)}}
    & \makecell[c]{\textbf{Peak VRAM}\\\textbf{(GiB)}}
    & \makecell[c]{\textbf{Disk storage}\\\textbf{(GiB)}} \\
    \midrule

    \multirow{2}{*}{ViT-B/32, 8-task}
    & MASS & 102.16 & 0.80 & 5.97 & 13.46 \\
    & MATE & 72.54 & 2.95 & 6.43 & 13.46 \\
    \midrule

    \multirow{2}{*}{ViT-B/16, 20-task}
    & MASS & 72.06 & 1.03 & 12.21 & 19.58 \\
    & MATE & 64.46 & 3.60 & 11.83 & 19.58 \\
    \midrule

    \multirow{2}{*}{ViT-L/14, 20-task}
    & MASS & 149.27 & 1.08 & 38.33 & 59.98 \\
    & MATE & 143.29 & 4.60 & 38.41 & 59.98 \\
    \bottomrule
\end{tabular}
}
\end{table}

\paragraph{\pdfdiff{Computational overhead.}}
\pdfdiff{
MATE collects representations from multiple layers during
the first backbone forward pass, retaining the same two
forward passes as MASS. As expected, multi-layer aggregation increases routing time by 2.15–3.52 ms.
Interestingly, despite the additional routing cost, MATE
achieves lower mean end-to-end inference latency in all
three configurations, with reductions of 29.0\%, 10.5\%,
and 4.0\%, respectively.
Stage-wise measurements indicate that dynamic TSV-M merging
dominates the average inference time.
Since the two methods may select different candidate task
sets, differences in merging costs and cache reuse can
outweigh the additional routing overhead.
This suggests that multi-layer routing does not necessarily increase overall online inference latency and can even yield faster end-to-end inference under the evaluated settings.
\par
}

\paragraph{\pdfdiff{Storage and memory overhead.}}
\pdfdiff{
MATE reuses the existing SVD/TSV representations and requires
no additional persistent model storage.
Peak GPU memory remains comparable between the two methods, with differences also affected by the cache of dynamically merged models. Although SVD representations and dynamic model caching remain substantial costs inherited from MASS, MATE introduces no additional persistent model storage. 
\par
}

\end{document}